\documentclass{article}

\usepackage[preprint]{neurips_2026}

\makeatletter
\def\blfootnote{\xdef\@thefnmark{}\@footnotetext}
\makeatother

\usepackage[utf8]{inputenc} 
\usepackage[T1]{fontenc}    
\usepackage{hyperref}       
\usepackage{url}            
\usepackage{booktabs}       
\usepackage{amsmath}        
\usepackage{amsfonts}       
\usepackage{nicefrac}       
\usepackage{microtype}      
\usepackage{xcolor}         
\usepackage{tikz}           
\usetikzlibrary{arrows.meta,positioning}
\usepackage{makecell} 
\usepackage{graphicx} 
\title{What Frozen VLAs Already Know About Success:\\A Probing Study of Value-Like Structure in Foundation Robot Policies}

\author{%
  \textnormal{Jiachen Zhang$^{1,2,*}$, Junnan Nie$^{1,*}$, Junyi Lao$^{1,*}$,}\\ 
  \textnormal{Wei Cheng$^{1}$, Chenghao Liu$^{1}$, Jiaxin Jiang$^{2}$, Songfang Huang$^{1,\dagger}$} \\
}

\begin{document}
\maketitle
\blfootnote{\hspace{-1.8em}\textsuperscript{*}Equal contribution. \textsuperscript{$\dagger$}Corresponding author.}
  \blfootnote{\hspace{-1.8em}$^1$Peking University. $^2$China Agricultural University.}
  \blfootnote{\hspace{-1.8em}Emails: \texttt{z89498323286@gmail.com, 2501213203@stu.pku.edu.cn, jylao25@stu.pku.edu.cn,}\\
  \texttt{ weicheng25@stu.pku.edu.cn, 2401213202@stu.pku.edu.cn, nongdamanong@cau.edu.cn, hsf@pku.edu.cn}}
\begin{abstract}
  Vision--language--action (VLA) policies are trained to imitate actions; their loss never asks them to estimate reward, progress, or future success.  Their frozen representations nevertheless carry such information, and it can be read out and used to guide action choice without retraining the policy.  From mixed successful and failed manipulation trajectories on LIBERO-Goal, we recover Monte-Carlo outcome targets using lightweight linear probes on frozen features.  The targets are consistently predictable from OpenVLA, Pi0.5, DINOv2, and CLIP features, and substantially less so from baselines built on progress, time-to-go, task identity, or proprioception.  To rule out task and temporal shortcuts, we evaluate the probes under same-task, same-timestep matched comparisons: Pi0.5 probes still reach roughly 92\% pairwise ordering accuracy, while label-shuffled controls stay at chance.  Used as a test-time selector over sampled Pi0.5 action prefixes, the same probe turns this offline finding into behavior: on push-plate, success rises from 26.7\% under greedy decoding to 44.3\%, with a second positive case on wine-rack.  The gains are not universal and require additional inference compute, but the underlying finding is clean: frozen VLAs already encode information about success that their imitation objective never explicitly demands.
\end{abstract}

\begin{figure}[t]
  \centering
  \includegraphics[width=\linewidth]{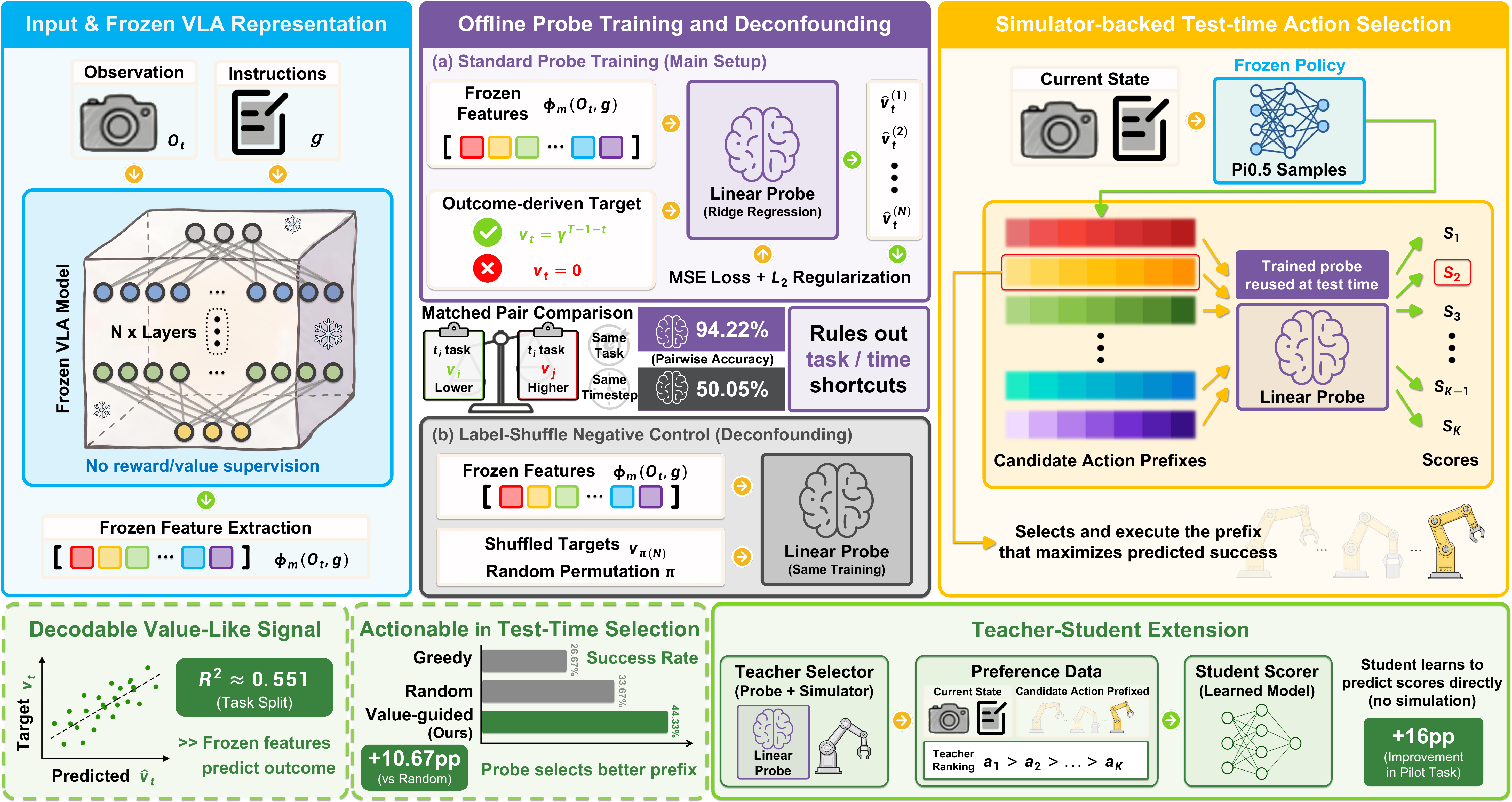}
  \caption{Overview of the probing-to-selection pipeline.  Frozen VLA and visual representations are probed for value-like signal on offline trajectories, then tested under same-task, same-timestep matched controls that hold task identity and elapsed time fixed by construction.  The selected probe is used as a simulator-backed evaluator over sampled Pi0.5 action-prefix candidates, turning the offline readout into a test-time selection signal without updating the policy.}
  \label{fig:framework-overview}
\end{figure}

\section{Introduction}
\label{sec:intro}

A vision--language--action (VLA) policy is trained to produce actions, following the imitation-driven recipe used by recent robot foundation policies \citep{brohan2022rt1,brohan2023rt2,kim2024openvla,black2024pi0,physicalintelligence2025pi05}.  It is never asked to estimate whether a state is close to success, whether progress has been made, or whether one candidate action is more promising than another.  Yet the trajectories one collects from such policies in deployment---successful and failed rollouts, together with the outcome labels they naturally carry---contain exactly this kind of information.  The question has two parts.  Does any of it survive inside the frozen representation, in a form a linear probe can recover?  And if it does, can that same readout alter which action the policy takes?

If such structure is present, it changes what a VLA representation is taken to be.  More than a conditional action generator, it would also carry outcome-relevant state information that is present without supervision for evaluation.  This in turn would open a route to influencing decisions at test time without any policy update, by ranking candidate actions through a representation-derived score, in contrast to selection methods that require a separately trained reward, cost, or predictive model \citep{ebert2018visual,chi2023diffusion}.  The opposite outcome would be just as informative.  If the signal is absent, or survives only as a dataset artifact, then current VLA representations are bounded by the imitation objective they were trained on.  Test-time improvements, under that scenario, have to come from external modules rather than from the policy itself.

The fact that training data contain outcome information does not, by itself, mean that frozen representations preserve it in a useful form.  Manipulation data are full of surface-level shortcuts a probe could exploit instead.  Tasks have different baseline success rates, so a probe could appear to predict outcome by latching onto task identity.  Timestep is correlated with progress, so a probe could appear to predict outcome by reading off elapsed time.  Successful and failed rollouts also differ in coarse visual statistics, which a probe might pick up without ever encoding anything action-relevant.  We therefore design controls that target each of these explanations directly, and treat the question of representation content as separate from the question of behavioral consequence.

We look for the signal in three places: whether it can be decoded from frozen features at all, whether it survives the hardest shortcut control we can construct, and whether it changes behavior when used to rank candidate actions at test time.

The signal is there, and the evidence falls into three consistent pieces.  Linear probes \citep{alain2016linear,hupkes2018diagnostic} on frozen Pi0.5, OpenVLA, DINOv2, and CLIP features \citep{kim2024openvla,physicalintelligence2025pi05,oquab2023dinov2,radford2021clip} predict Monte-Carlo outcome targets far better than progress, time-to-go, task identity, or proprioception, with the breadth across families arguing against a single-model artifact.  The signal also survives a same-task, same-timestep matched control: after fixing both task and timestep so that obvious shortcuts no longer separate the data, Pi0.5 probes still reach roughly 92\% pairwise ordering accuracy, while label-shuffled controls collapse to chance.  Most importantly, the same probe can rank Pi0.5's own candidate action prefixes at test time and change which one is executed: on push-plate, success rises from 26.7\% under greedy decoding to 44.3\%, with a second positive case on wine-rack.  The readout that decoded outcome offline now alters which action is taken online, on the very tasks where the underlying policy already struggles.

This paper therefore makes three claims about frozen VLA representations:

\begin{itemize}
    \item \textbf{A value-like signal is decodable from multiple VLA backbones}, rather than being an artifact of a single model or layer.  Several frozen representation families---VLA, VLM, and general-purpose visual encoders---carry it at comparable strength, while scalar nuisance baselines do not.
    \item \textbf{The value-like signal is not a task or temporal shortcut}, under a stringent matched-pair control.  After fixing both task and timestep, probes still order outcome-labelled pairs correctly, while label-shuffled controls collapse to chance.
    \item \textbf{The value-like signal is behaviorally usable}: used as a test-time ranker over Pi0.5 action prefixes, it changes which prefix is executed on hard manipulation tasks, at a measurable compute cost.
\end{itemize}

\section{Related Work}
\label{sec:related}

\paragraph{VLA policies and the imitation objective.}
Recent robot foundation models treat manipulation as conditional sequence generation from images, language, and robot state.  RT-1, RT-2, and PaLM-E established that Transformer policies could scale on real-robot demonstrations and exploit web-scale pretraining \citep{brohan2022rt1,brohan2023rt2,driess2023palme}.  Subsequent open and generalist policies---Open X-Embodiment, OpenVLA, OpenVLA-OFT, Octo, SmolVLA, \(\pi_0\), and \(\pi_{0.5}\)---pushed this line further by aggregating cross-embodiment data, releasing open checkpoints, improving fine-tuning efficiency, and exploring flow-based action generation \citep{openx2023,kim2024openvla,kim2025openvlaoft,octo2024,shukor2025smolvla,black2024pi0,physicalintelligence2025pi05}.  All of these policies are trained to imitate actions; none are trained to estimate reward, progress, or future success.  The representational consequences of that asymmetry remain comparatively underexplored, and that asymmetry is the starting point for our investigation.  Our experiments use OpenVLA and Pi0.5 as primary VLA backbones, with LIBERO-Goal \citep{liu2023libero} as the main benchmark and CALVIN \citep{mees2022calvin} together with general-purpose visual encoders such as CLIP and DINOv2 \citep{radford2021clip,oquab2023dinov2} as cross-checks.

\paragraph{Probing and value functions.}
Linear probes and diagnostic classifiers measure what information is accessible from hidden representations \citep{alain2016linear,hupkes2018diagnostic}, and have been used across language and vision representations to identify structure not predicted by the training objective.  Our use of probes follows this diagnostic tradition but targets an embodied control question.  In reinforcement learning, a value function has a Bellman-consistent meaning tied to expected return under a policy \citep{sutton2018reinforcement}, and is typically learned through bootstrapped updates rather than direct outcome regression.  Our targets are deliberately weaker: Monte-Carlo outcome labels constructed from completed manipulation trajectories.  We therefore refer to the readout as a \emph{value-like} signal, and our claims do not assume Bellman consistency.  Two further differences matter.  We do not stop at offline decodability; we add same-task, same-timestep matched controls that explicitly remove the shortcuts most often left implicit, and we use the same readout inside a test-time action-selection loop, so that representation content is judged not only by what a probe can fit but also by what it can change.

\paragraph{Test-time selection without a learned reward model.}
Sampling-based control and visual model-predictive control choose actions by evaluating candidate futures \citep{ebert2018visual}, and Diffusion Policy and related receding-horizon visuomotor policies place candidate generation at the center of robotic control \citep{chi2023diffusion}.  These approaches typically rely on an external reward, cost, or world model to score candidates.  Our setting differs in a single but consequential way: the learned scoring term comes from a probe trained on the policy's own frozen features, with no policy update and no separately trained reward model.  The selector is therefore a diagnostic device---a way to test whether a ranking signal already internal to the policy can help choose prefixes from the policy's own samples.

\section{Method: A Probing-to-Selection Protocol}
\label{sec:method}

We use a staged probing-to-selection protocol throughout.  Its purpose is to make the three claims falsifiable rather than to be evaluated as a methodological contribution in its own right.  Given an existing policy or encoder, the protocol extracts frozen features, fits standardized linear probes on offline trajectories, evaluates the probes under matched nuisance controls, and uses the same readout to select among candidate action prefixes at test time.  Policy parameters are fixed throughout.

Three requirements shape the staging.  At the offline stage the probe has to be read without loopholes, which is why we use standardized linear ridges and report demo and task splits side by side, without feeding task identity, progress, or elapsed time in as covariates---these enter only as separate baseline rows.  The matched-pair stage then exists because offline \(R^2\) on its own cannot distinguish genuine within-task ordering from the kind of task or timestep shortcut a probe could exploit.  We therefore hold both nuisances fixed by construction rather than treating them as controls after the fact.  The third stage closes the loop: the same readout, with no retraining, is plugged into a candidate-selection loop on top of the policy's own samples, so that whatever the probe has learned is judged not only by what it can fit but also by what it can change.  Failure at any stage is informative in its own right, and we record such failures rather than smoothing them out.

\subsection{Problem Setup and Value-Like Signals}
\label{sec:method-setup}

We consider language-conditioned manipulation episodes.  At timestep \(t\), the policy observes an image-based observation \(o_t\), a language instruction \(g\), and proprioceptive state when available, and produces an action or an action chunk.  Each trajectory has a binary task outcome.  For a successful trajectory of length \(T\), we assign a discounted Monte-Carlo-style target
\[
v_t = \gamma^{T-1-t}, \qquad \gamma = 0.99,
\]
so later states in a successful trajectory receive larger targets.  For failed trajectories, \(v_t=0\) for all timesteps.  We call this a value-like target rather than an environment value function: it is derived from observed trajectory outcomes and episode time, and is used to probe whether frozen representations contain information correlated with eventual success.  We additionally store scalar progress \(t/T\), time-to-go \((T-t)/T\), and task identity as nuisance variables and baselines.

Two design choices are worth naming explicitly.  We set \(\gamma=0.99\), which gives an effective horizon of roughly 100 steps---appropriate for LIBERO-Goal episodes that run to at most 300 steps with most successful trajectories completing in under 150.  We use Monte-Carlo returns constructed from completed trajectories rather than bootstrapped value estimates, which removes the need for a separately learned value network and avoids the instabilities that come with bootstrapping on frozen features that were never trained for prediction consistency.  The cost is that the target is noisier than a smoothed TD estimate; the benefit is that the probe's success or failure reflects genuine representation content rather than a secondary learning artifact.

\subsection{Offline Probing of Frozen Representations}
\label{sec:method-probing}

Let \(\phi_m(o_t,g)\in\mathbb{R}^d\) denote a frozen representation extracted from model or feature family \(m\).  The feature families include VLA/VLM layers, vision encoders, proprioception, random projections, and scalar nuisance features.  For each feature family, we fit a linear ridge probe
\[
\hat v_t = w^\top \tilde{\phi}_m(o_t,g) + b,
\]
where \(\tilde{\phi}\) is standardized using training-set statistics.  The ridge penalty is selected by cross-validation over a fixed grid of regularization strengths.  We report \(R^2\), Spearman correlation, RMSE/MAE, and per-task \(R^2\) on held-out examples.

We use two split regimes.  In the demo split, trajectories are grouped by suite, task, and source demonstration; train/test groups are sampled within each task.  This tests whether a representation captures value-like structure across held-out trajectories from seen tasks.  In the task split, groups are defined by suite and task, so held-out examples come from unseen tasks.  This is a stricter test and is treated separately throughout the paper.  Task identity, progress, and time-to-go are not used as privileged covariates for the learned VLA probes; they are trained as separate baselines.

\subsection{Deconfounding with Matched Controls}
\label{sec:method-controls}

Offline probing can be inflated by nuisance structure.  We therefore use a same-step matched-pair control.  Given a trained probe and a feature table, we group rows by task name and timestep.  Within each group, we form low/high pairs whose value-like labels differ by at least 0.20.  A probe is correct on a pair if the higher-label row receives the higher probe score; ties receive half credit.  The resulting metric is
\[
\frac{1}{|\mathcal{P}|}\sum_{(i,j)\in\mathcal{P}}
\left(\mathbf{1}\{\hat v_j > \hat v_i\} + \frac{1}{2}\mathbf{1}\{\hat v_j = \hat v_i\}\right),
\]
where \(v_j>v_i\) and both rows share the same task and timestep.  We also compute a label-shuffle control by randomly swapping which row is treated as the positive member of each pair.  This control tests whether the probe contains within-task, within-timestep ordering information, rather than merely encoding task ID or elapsed time.

\subsection{Test-Time Candidate Selection}
\label{sec:method-selection}

To test actionability, we use a best-of-\(K\) selection procedure with the frozen Pi0.5 policy.  At each replan step, the runtime samples \(K=16\) action chunks using deterministic candidate seeds derived from the episode seed, replan index, and candidate index.  We execute only a short prefix of each chunk; in the main experiments the prefix length is five actions.  We picked \(K=16\) because an early two-task sweep on open middle and open top drawer showed overall success climbing from 60\% at \(K{=}4\) and \(K{=}8\) to 80\% at \(K{=}16\): the candidate set needed enough breadth to contain a meaningfully better prefix before the probe could discriminate one.  We treat this sweep as design evidence rather than as formal validation, and report the full numbers in Appendix~\ref{sec:appendix-boundary}.  The five-action prefix is short enough that the simulator-evaluated rollout is dominated by near-term commitment rather than drift, but long enough that the resulting observation has actually moved.  Greedy decoding executes the policy's decoded chunk directly.  Random selection samples the same candidate set but chooses a prefix uniformly at random.

Value-guided selection uses the learned probe as one component of a simulator-backed teacher selector.  For each candidate prefix \(a_{t:t+h}^{(k)}\), we restore the simulator to the current snapshot, roll out the prefix, and record three quantities: whether the short rollout reaches task success, the accumulated simulator reward \(r_k\), and the probe score \(s_k=\hat v(\phi(o_{t+h}^{(k)},g))\) extracted from the resulting observation.  In the formal Phase-4 runs, the selector first restricts to successful candidate prefixes if any are found and then chooses the highest-scoring one; if no prefix succeeds during the short rollout, it chooses by normalized reward plus normalized probe score:
\[
k^\star =
\begin{cases}
\operatorname*{arg\,max}_{k\in\mathcal{S}} s_k, & \text{if } \mathcal{S}\neq\emptyset,\\
\operatorname*{arg\,max}_{k\in\{1,\ldots,K\}} \left[z(r_k)+z(s_k)\right], & \text{otherwise},
\end{cases}
\]
where \(\mathcal{S}\) is the set of candidate prefixes that succeed in the simulator rollout and \(z(\cdot)\) denotes normalization within the current candidate set.  After selection, the simulator is restored to the original snapshot and only the chosen prefix is committed to the episode.  The procedure uses simulator access to expose candidate futures, so it is best read as a diagnostic instrument for testing whether the decoded signal can guide action choice when the search budget is paid in full, rather than as a runtime controller.

\subsection{Scope of the Protocol}
\label{sec:method-scope}

The three stages are deliberately kept separate.  Decodability asks whether frozen features predict a value-like target offline.  The matched-pair stage asks whether that prediction survives once task and timestep shortcuts are held fixed.  Online selection then asks whether the same readout can pick a better action prefix at test time, on top of the policy's own samples.  A positive result at one stage motivates, but does not guarantee, a positive result at the next, and localizing where the signal stops being useful is itself a finding.  We therefore never collapse the three into a single headline number, and we preserve the per-stage outcome whether it is positive, borderline, or null.

Several constraints follow from this design.  The probe target is outcome-derived, not a Bellman-consistent value estimate, and our claims do not assume otherwise.  The test-time selector relies on simulator rollouts to evaluate candidates, which is the source of the additional wall-clock cost reported in Section~\ref{sec:rq3-online} and the reason selection helps only when the sampled candidate set actually contains alternatives that differ in downstream outcome.

\section{Experiments}
\label{sec:experiments}

The experiments follow the three claims laid out in the introduction.  Section~\ref{sec:rq1-decodability} addresses decodability and the matched-control evidence together, because the second is meaningless without the first.  Section~\ref{sec:rq3-online} then asks whether the same readout, inserted into a candidate-selection loop, actually changes which action Pi0.5 executes.  A distinction runs through both: formal evidence uses balanced production runs with matched strategy comparisons, whereas supporting evidence reports earlier or exploratory runs that point in the same direction without sharing the balanced design.  We never pool the two, and the appendix retains the supporting runs with their original status.

\begin{table}[t]
\centering
\setlength{\tabcolsep}{4pt}
\begin{tabular*}{\linewidth}{@{\extracolsep{\fill}}l l c l r@{}}
\toprule
Feature source & Demo best feature & Demo \(R^2\) & Task best feature & Task \(R^2\) \\
\midrule
Pi0.5 & vision encoder & 0.7377 & vision encoder & \textbf{0.5510} \\
OpenVLA-OFT & LLM layer 07 & \textbf{0.7561} & vision backbone & 0.5505 \\
OpenVLA & LLM layer 07 & 0.7553 & vision backbone & 0.5493 \\
OpenVLA-v0.1 & LLM layer 20 & 0.7354 & LLM layer 20 & 0.5429 \\
SmolVLA & VLM final & 0.7028 & VLM layer 16 & 0.5257 \\
DINOv2 & DINOv2 & 0.6899 & DINOv2 & 0.5104 \\
CLIP & CLIP & 0.6747 & CLIP & 0.5095 \\
Random projection & Random projection & 0.5568 & Random projection & 0.3916 \\
Proprioception & Proprioception & 0.2010 & Proprioception & 0.1107 \\
Pi0 & VLM final & 0.3040 & vision encoder & 0.0702 \\
Progress & Progress & 0.0325 & Progress & 0.0302 \\
Time-to-go & Time-to-go & 0.0325 & Time-to-go & 0.0302 \\
Task one-hot & Task one-hot & -0.0009 & Task one-hot & -0.0018 \\
\bottomrule
\end{tabular*}
\caption{LIBERO-Goal offline value probing across frozen representation families.  All rows use the same 311,719-row mixed dataset and report the best feature or layer within each family; rows are ordered by task-split \(R^2\).}
\label{tab:representation-evidence}
\end{table}

\subsection{Experimental Setup}
\label{sec:experimental-setup}

We instantiate the protocol on LIBERO-Goal as the main benchmark.  Offline probing uses 311,719 frame-level rows from 1,400 mixed successful and failed trajectories.  CALVIN-D and RobotWin provide cross-benchmark representation checks.  The feature set includes OpenVLA and OpenVLA-OFT layers, Pi0.5 vision/VLM features, SmolVLA features, CLIP, DINOv2, proprioception, random projections, scalar progress/time features, and task one-hot features.

The online experiments use the official Pi0.5 LIBERO policy.  Unless otherwise specified, random and value-guided selection use \(K=16\) candidate chunks and execute a five-action prefix, while greedy executes the decoded chunk directly.  The value-guided selector uses the selected Pi0.5 \texttt{vlm\_final\_output} probe trained on \texttt{value\_mc} with the demo split as the learned score inside the simulator-backed teacher rule defined in Section~\ref{sec:method-selection}.  Episodes are capped at 300 environment steps.  The main online analysis is restricted to the formal merged Phase-4 runs: the balanced hard-3 run, the push-plate balanced follow-up, and the wine-rack balanced follow-up.  Smoke runs, aborted runs, and earlier key-fix runs are excluded.

For absolute online success rates, Wilson 95\% confidence intervals are computed in the analysis pipeline; the main table reports successes and point estimates to keep the task-level comparison readable.  Strategy contrasts are computed on paired units sharing the same run family, server, task, and episode index; the analysis pipeline computes paired bootstrap 95\% intervals and exact McNemar \(p\)-values, with Fisher exact tests as an unpaired sensitivity check.  The main table reports the McNemar test for the value-guided versus random contrast.  This separation keeps the exploratory and confirmatory roles of the experiments distinct: early unbalanced runs are useful for design, while the claims below are based on paired or balanced production runs.

We use the term \emph{hard non-ceiling} to label tasks where greedy decoding succeeds at most about 40\% of the time, so that a selection strategy has measurable headroom but the task is not pathological.  Under this rule, the merged Phase-4 hard-3 set---open middle drawer (38.7\% greedy), wine rack (35.7\%), and push plate (26.7\%)---qualifies, while the bowl-on-stove and bowl-on-cabinet tasks at \(\geq\)98\% greedy are ceiling cases for which selection cannot meaningfully help and are excluded by construction.

For reproducibility, each linear ridge probe is trained over five random seeds with a fixed grid of regularization strengths, and the same-step controls are reported across ten matched runs spanning two probe targets and five seeds each.  Online rollouts are distributed across three independent servers to remove single-machine confounds, with seed assignment fixed in advance per server (seeds 1111--3333 for hard-3, 4444--6666 for the push-plate follow-up, 7777--9999 for the wine-rack follow-up).  Probes train in under five minutes on a single GPU; the dominant cost in the pipeline is the simulator-backed candidate evaluation reported in Section~\ref{sec:rq3-online} and Appendix~\ref{sec:appendix-compute}.

\subsection{Representation Evidence and Deconfounding}
\label{sec:rq1-decodability}

Table~\ref{tab:representation-evidence} reports the offline probing comparison on LIBERO-Goal.  Each row uses the same convention: the best feature or layer within one representation family, evaluated by demo-split and task-split \(R^2\) on the same 311,719-row mixed dataset.  OpenVLA-OFT, OpenVLA, OpenVLA-v0.1, Pi0.5, SmolVLA, DINOv2, and CLIP all sit substantially above proprioception and the scalar nuisance features.  The scalar baselines themselves deserve a closer look: progress and time-to-go each reach \(R^2 \approx 0.03\), and task identity is essentially zero.  An offline probe that achieves task-split \(R^2 \approx 0.55\) on a frozen VLA is therefore not picking up something a stopwatch or a task tag would have captured for free.  Random projections are nontrivial at \(R^2{=}0.39\) on the task split, which sharpens rather than weakens the comparison: part of the target is recoverable from dataset geometry alone, and the learned VLA and visual features still sit clearly above that floor.

Breadth across families is what turns the result into a representation-level statement rather than a one-model curiosity.  Four families with different training recipes---autoregressive VLA (OpenVLA, OpenVLA-OFT), VLA-distilled VLM (Pi0.5, SmolVLA), self-supervised vision (DINOv2), and contrastive vision-language (CLIP)---all surface a value-like signal at comparable strength on the same data.  The exception is informative.  Pi0, a flow-matching action head built on top of a VLM, lands at task-split \(R^2{=}0.0702\), close to the scalar baselines: a representation that is heavily specialized for action generation does not appear to retain the same outcome structure, even though Pi0.5 in the same lineage does.  One possible interpretation, consistent with the rest of the table, is that the value-like signal lives in the parts of the stack that still encode goal-conditioned scene state, and is squeezed out where the representation has been compressed toward action-token prediction.  As a cross-benchmark sanity check, OpenVLA on CALVIN-D demo split reaches \(R^2{=}0.3007\) (Spearman 0.6034, Table~\ref{tab:appendix-cross-benchmark}); the signal is weaker outside LIBERO-Goal but is not LIBERO-only.

Shortcut controls answer a different question, so we keep them in a separate table rather than folding them into the \(R^2\) comparison.  Table~\ref{tab:same-step-controls} fixes task and timestep, then asks whether the probe can still order high-value and low-value rows.  The primary Pi0.5 probe reaches 94.22\% pairwise accuracy while label shuffle sits at 50.05\%; the per-task numbers on drawer, push-plate, and wine-rack are similarly high.  Across ten same-step runs spanning two feature choices and multiple seeds, mean pairwise accuracy is 92.16\% and shuffle stays at 49.67\%.  This does not turn the probe into a Bellman value function, but it does rule out the task-identity and elapsed-time explanations that a probe might otherwise be exploiting.

The robustness of this gap is itself part of the result.  The ten same-step runs cover both \texttt{vlm\_final\_output} and \texttt{vlm\_layer\_08} as feature choices, and within each feature choice we trained five independent probe seeds.  The pairwise accuracy across the ten runs ranges from 89.58\% to 94.22\%, while the corresponding label-shuffle controls stay tightly between 49.01\% and 50.05\%.  No probe configuration drops below 89.58\%, and no shuffle configuration rises above 51.19\%.  The matched-pair signal is therefore not an artifact of a single probe seed, a single feature layer, or a single training run; it is reproducible under the most aggressive within-task, within-timestep control we know how to construct.

\begin{table}[t]
\centering
\setlength{\tabcolsep}{4pt}
\begin{tabular*}{\linewidth}{@{\extracolsep{\fill}}l l r r c c@{}}
\toprule
Probe set & Scope & Pairs/runs & Pair acc. & Shuffle acc. & Gap \\
\midrule
Primary probe & Overall & 4,605 pairs & \textbf{94.22\%} & 50.05\% & \textbf{+44.17 pp} \\
Seed/layer robustness & Overall & \(10\times4{,}605\) pairs & 92.16\% & 49.67\% & +42.49 pp \\
Primary probe & Open middle drawer & 1,053 pairs & 92.31\% & 51.19\% & +41.12 pp \\
Primary probe & Push plate & 1,892 pairs & 94.56\% & 50.79\% & +43.77 pp \\
Primary probe & Wine rack & 1,660 pairs & 95.06\% & 48.49\% & +46.57 pp \\
\bottomrule
\end{tabular*}
\caption{Same-step controls against task and timestep shortcuts.  Each pair shares the same task and timestep; shuffle accuracy is the label-shuffle negative control.}
\label{tab:same-step-controls}
\end{table}

\subsection{Online Positive Evidence}
\label{sec:rq3-online}

Having established decodability and the matched-control evidence, we now ask whether the decoded signal can change behavior.  Table~\ref{tab:online-positive} reports the formal Pi0.5 online comparisons.  Early pilots, OpenVLA confirmation runs, and the \(K\) and temperature sweeps are useful supporting evidence but do not share the same balanced production design, so we keep them out of this table; they appear in Appendix~\ref{sec:appendix-boundary} with their original status preserved.

On the pooled hard-3 suite, value-guided selection reaches 42.44\% success against 36.89\% for random and 31.11\% for greedy.  The gain over greedy is large and clean; the gain over random is smaller and lands at \(p{=}0.0614\) on the suite-level McNemar contrast, and we treat this contrast as borderline rather than significant.  The aggregate hides the per-task structure, and the per-task structure is where the actionability story actually sits.

The clearest case is push-plate.  On 300 paired episodes, value-guided selection improves over random by 10.67 percentage points and over greedy by 17.67 points, with the McNemar contrast against random at \(p{=}0.003\).  The policy's own greedy decoding succeeds only 26.7\% of the time here; value-guided selection nearly doubles that, reaching 44.3\%.  Because the candidate set is sampled from the same Pi0.5 policy, this is not a case of an external expert injecting better trajectories.  The better continuations were already inside the sampling distribution; what the probe does is reliably find them.

Wine rack behaves the same way with smaller margins.  Value-guided beats random by 8.00 and greedy by 8.33 points, with McNemar \(p{=}0.026\).  On its own this would be a single positive task, easy to explain away as noise; pooled with push plate under an identical balanced design, it becomes the second independent confirmation that the probe is doing something systematic.

Open middle drawer then marks the boundary.  Value-guided selection moves only \(+2.00\) against random and \(+0.67\) against greedy, well within sampling noise; we call this a no-gain case and report it as such.  The full numbers live in Appendix~\ref{sec:appendix-boundary}.  The boundary is useful because it rules out the simpler explanation that selection always helps when candidates are available: here the candidates are available, the probe is the same, the policy is the same, and the gain is gone.  What differs is the structure of the task itself, and locating that boundary is part of what the staged design is for.

\begin{table}[t]
\centering
\setlength{\tabcolsep}{2pt}
\begin{tabular}{@{}
p{0.21\linewidth}
p{0.1\linewidth}
p{0.1\linewidth}
>{\centering\arraybackslash}p{0.1\linewidth}
>{\centering\arraybackslash}p{0.1\linewidth}
>{\centering\arraybackslash}p{0.1\linewidth}
>{\centering\arraybackslash}p{0.1\linewidth}
>{\centering\arraybackslash}p{0.12\linewidth}
@{}}
\toprule
Run / task & Greedy & Random & Value & \(\Delta\) V--R & \(\Delta\) V--G & \raisebox{2pt}{$p_{\scriptscriptstyle\mathrm{VR}}$} & Wall V/R \\
\midrule
Hard-3 aggregate & 140/450 (31.1\%) & 166/450 (36.9\%) & \textbf{191/450 (42.4\%)} & +5.56 & +11.33 & 0.0614 & 2.06\(\times\) \\
Push plate merged & 80/300 (26.7\%) & 101/300 (33.7\%) & \textbf{133/300 (44.3\%)} & \textbf{+10.67} & \textbf{+17.67} & 0.00294 & 2.10\(\times\) \\
Wine rack merged & 107/300 (35.7\%) & 108/300 (36.0\%) & \textbf{132/300 (44.0\%)} & +8.00 & +8.33 & 0.0264 & 2.01\(\times\) \\
\bottomrule
\end{tabular}
\caption{Formal online evidence that probe-guided selection improves Pi0.5 on selected hard tasks, at a compute cost.  Strategy columns report successes/episodes with success rate in parentheses; \(p_{\mathrm{VR}}\) is the paired McNemar exact \(p\)-value for value-guided versus random selection.}
\label{tab:online-positive}
\end{table}

The pattern is positive but selective, and the improvement is bought with extra inference compute.  In the formal Pi0.5 runs, value-guided selection takes roughly twice as long per episode as random selection and more than twenty times as long as greedy decoding (Appendix~\ref{sec:appendix-compute}).  In wall-time-per-extra-success terms, push plate costs about 39 minutes of additional compute per additional success over random, wine rack about 47 minutes; drawer, where there are essentially no extra successes, costs about 185 minutes.  This cost is central to the interpretation of the result.  What the result establishes is narrower and more specific: a frozen, representation-derived score can be converted into demonstrably better choices when the search budget is paid in full.  Whether this can be turned into a deployment-grade procedure---through cheaper candidate evaluation, learned scorers, or compute-matched comparison---is a separate engineering problem that we do not solve here.

\subsection{Goal-Conditioning Diagnostics}
\label{sec:goal-conditioning}

The same-step matched-pair control rules out task and timestep shortcuts, but it does not by itself say what the probe is reading.  A natural worry is that the probe relies on visual state alone and treats the language goal as decoration.  Goal-swap controls address this directly: they substitute the language instruction while holding the visual observation fixed, then ask whether the probe score moves.  A vision-only run, where the goal embedding is bypassed by construction, produces exactly zero feature change and exactly zero score change across 14,080 pairs.  This negative control matters more than it first appears.  It guarantees that any non-zero shift we observe downstream of a real goal swap can only come from the language pathway, not from numerical drift in the rest of the pipeline.

Pi0.5 VLM features do move under language swaps.  Mean feature \(L_2\) shifts by 1.96 with cosine similarity \(0.9985\), confirming that the VLM stack reads the goal text rather than ignoring it, while the score itself moves by a mean delta of \(+0.12\) on the seed-7 \texttt{vlm\_final\_output} probe with the original goal scoring higher on 53.7\% of pairs.  A second seed and a deeper VLM layer give means of \(-0.08\) and \(-0.02\), so the sign is not uniformly aligned with the original goal: drawer-style tasks tend to score lower on the original goal, while rack and container tasks score higher (Appendix~\ref{sec:appendix-goal-swap}).  One reading of this asymmetry is that the probe inherits a goal-conditioned representation from the VLM stack but does not behave as a calibrated reward model that always prefers the matched goal.  The asymmetric direction does not weaken the same-step matched-pair finding, which already holds task identity fixed by construction and therefore cannot be inflated by whatever the language pathway is encoding asymmetrically.  Per-task numbers, \(K\) and temperature sweeps, additional boundary tasks, and a small teacher-to-student experiment appear in Appendix~\ref{sec:appendix-boundary}.

\subsection{Summary of Findings}

Outcome-derived value-like targets are decodable from a broad set of frozen representations on LIBERO-Goal, and the decoding survives a same-task, same-timestep matched control that defeats the most natural shortcut explanations.  Plugged into Pi0.5's candidate-selection loop, the same probe lifts push-plate and wine-rack success on hard non-ceiling tasks while leaving drawer near unchanged.  The improvement is real and bought with compute.  Supporting OpenVLA, \(K\)-sweep, and temperature results---together with the boundary cases and the goal-swap per-task breakdown---are reported in Appendix~\ref{sec:appendix-boundary}.

Read across the three stages, a consistent picture emerges.  The offline table says the value-like signal is there at representation-level strength.  The matched-pair table says it is not an artifact of task identity or elapsed time.  The online table says the same readout changes Pi0.5's behavior on exactly the tasks where the baseline policy is weak enough to leave room; where the policy already succeeds, or fails in a way candidate diversity cannot fix, the gain collapses.  The pattern across the three stages is consistent with a frozen representation that encodes a value-like signal the policy itself does not fully exploit during decoding.

\section{Limitations}
\label{sec:limitations}
These results support a deliberately narrow interpretation. We do not claim a Bellman-consistent value function inside VLA models, the probe target is outcome-derived, and the online gains are task-dependent. The selector also relies on simulator-backed candidate evaluation: value-guided selection is roughly twice as expensive as random selection and more than an order of magnitude slower than greedy decoding. The method should therefore be read as a diagnostic intervention rather than a deployment-ready controller; making it practical would require cheaper candidate evaluation or a learned scorer that can replace simulator access.

\section{Conclusion}
\label{sec:conclusion}
A VLA policy is trained only to imitate actions, but its frozen representation already encodes more than that objective demands. Outcome-derived value-like targets are decodable from a wide range of frozen VLA, VLM, and visual encoders, with a strength that scalar progress, time-to-go, and task identity cannot account for; the decoding survives a same-task, same-timestep matched control that fixes the most natural shortcut explanations by construction; and the same readout, applied to Pi0.5's own sampled action prefixes at test time, raises success on push-plate from 26.7\% to 44.3\% and gives a second positive case on wine-rack while leaving drawer near unchanged. Three claims, three stages, and a single probe carries through all of them.

Taken together, the results suggest that an imitation-trained VLA is not only a black-box action generator: its frozen features already carry an outcome-related ordering that its own decoding does not fully exploit. Two questions follow naturally: what determines whether a sampled candidate set contains useful alternatives, and whether the simulator-backed evaluator can be replaced by a reliable learned scorer.

\bibliographystyle{plainnat}
\bibliography{references}


\appendix

\section{Additional Boundary and Ablation Results}
\label{sec:appendix-boundary}

This appendix reports auxiliary experiments that characterize where the probing-to-selection protocol transfers, where it becomes sensitive to the sampling regime, and the computational cost of simulator-backed candidate evaluation.  The main text uses only matched production runs for its formal online claims; the appendix keeps runs that differ in benchmark, split, sample size, or rollout design.  We use these results to characterize boundary conditions rather than to form another aggregate score.

\paragraph{Code-level online selector.}
The procedure below mirrors the Phase-4 Pi0.5 rollout implementation.  It is included because the online intervention depends on simulator state restoration: every candidate prefix is evaluated from the same snapshot, and the environment is restored again before the chosen prefix is committed.

\begin{center}
\begin{minipage}{0.96\linewidth}
\hrule
\vspace{4pt}
\textbf{Procedure 1. Simulator-backed Phase-4 selector.}

\textbf{Inputs:} frozen Pi0.5 runtime, current observation \(o_t\), instruction \(g\), simulator snapshot \(x_t\), trained probe \(\hat v\), feature extractor \(\phi\), candidate count \(K=16\), prefix length \(h=5\).
\begin{enumerate}
    \item Build the Pi0.5 model input from \(o_t\), \(g\), and snapshot metadata \(x_t\).
    \item For each candidate \(k=1,\ldots,K\):
    \begin{enumerate}
        \item Sample an action chunk from Pi0.5 using a deterministic seed derived from the episode seed, replan index, and candidate index.
        \item Keep the first \(h\) actions as the candidate prefix \(a_{t:t+h}^{(k)}\).
        \item Restore the simulator to \(x_t\), execute the prefix, and record success, accumulated reward \(r_k\), and the resulting observation.
        \item Extract the frozen feature \(\phi(o_{t+h}^{(k)},g)\) from the resulting observation and score it with the probe, \(s_k=\hat v(\phi(o_{t+h}^{(k)},g))\).
    \end{enumerate}
    \item If any candidate prefix succeeds during the simulator rollout, choose the successful candidate with the largest probe score \(s_k\).
    \item Otherwise, choose the candidate maximizing \(z(r_k)+z(s_k)\), where \(z(\cdot)\) normalizes values within the current candidate set.
    \item Restore the simulator to \(x_t\), enqueue the chosen prefix, and execute only that prefix in the real episode.
\end{enumerate}
\vspace{2pt}
\textbf{Baselines:} random selection samples the same \(K\) chunks and chooses one uniformly; greedy decoding executes the policy's decoded chunk without candidate evaluation.
\vspace{4pt}
\hrule
\end{minipage}
\end{center}

\paragraph{Task regimes and ceiling cases.}
The online experiments focus on tasks with enough headroom for selection to matter.  Table~\ref{tab:appendix-task-regimes} records the task regimes behind that choice.  The formal hard-task rows use the same merged Pi0.5 payloads as the main online analysis, while the high-success rows come from the earlier Pi0.5 greedy screening used to identify ceiling regimes.  Near-ceiling tasks are poor tests of a selector: when greedy decoding already solves almost every episode, a better ranking rule has little room to change the measured success rate.

\begin{table}[!htbp]
\centering
\setlength{\tabcolsep}{2pt}

\begin{tabular}{@{}
  >{\raggedright\arraybackslash}p{\dimexpr 0.23\linewidth - 1.5\tabcolsep\relax}
  >{\centering\arraybackslash}p{\dimexpr 0.18\linewidth - 1.5\tabcolsep\relax}
  >{\centering\arraybackslash}p{\dimexpr 0.12\linewidth - 1.5\tabcolsep\relax}
  >{\centering\arraybackslash}p{\dimexpr 0.48\linewidth - 1.5\tabcolsep\relax}
@{}}
\toprule
Task & Greedy success & Episodes & Role in the paper \\
\midrule
Open middle drawer & 38.67\% & 150 & hard non-ceiling boundary case \\
Push plate & 26.67\% & 300 & main positive online task \\
Wine rack & 35.67\% & 300 & secondary positive online task \\
Open top drawer & 82.00\% & 50 & early high-success screening task \\
Bowl on stove & 98.00\% & 50 & ceiling case, excluded from online claims \\
Bowl on cabinet & 100.00\% & 50 & ceiling case, excluded from online claims \\
\bottomrule
\end{tabular}
\caption{Task-regime evidence used to separate hard non-ceiling online tasks from near-ceiling tasks. Rows come from the formal merged Pi0.5 payloads or the earlier Pi0.5 greedy screening.}
\label{tab:appendix-task-regimes}
\end{table}

\paragraph{Cross-benchmark probing.}
Table~\ref{tab:appendix-cross-benchmark} asks whether the offline representation signal is confined to the main LIBERO-Goal setting.  CALVIN-D gives a positive demo-split check for OpenVLA, though its task split is much weaker.  RobotWin shows an even sharper split: Pi0.5 features give strong demo-split fits across seeds, but the task split has negative \(R^2\) despite weakly positive Spearman correlations.  We therefore use these rows as evidence for within-benchmark decodability beyond LIBERO, not as evidence that the current probes support robust cross-task transfer.

\begin{table}[!htbp]
\centering
\setlength{\tabcolsep}{4pt}
\begin{tabular}{@{}
  >{\raggedright\arraybackslash}p{\dimexpr 0.18\linewidth - 1.67\tabcolsep\relax}
  >{\centering\arraybackslash}p{\dimexpr 0.11\linewidth - 1.67\tabcolsep\relax}
  >{\centering\arraybackslash}p{\dimexpr 0.18\linewidth - 1.67\tabcolsep\relax}
  >{\centering\arraybackslash}p{\dimexpr 0.22\linewidth - 1.67\tabcolsep\relax}
  >{\centering\arraybackslash}p{\dimexpr 0.1\linewidth - 1.67\tabcolsep\relax}
  >{\centering\arraybackslash}p{\dimexpr 0.21\linewidth - 1.67\tabcolsep\relax}
@{}}
\toprule
Benchmark & Split & \(N\) & Feature & \(R^2\) & Extra statistic \\
\midrule
CALVIN-D & demo & 82,436 rows & OpenVLA L20 & 0.3007 & \(\rho=0.6034\) \\
CALVIN-D & task & 82,436 rows & OpenVLA L07 & 0.0393 & \(\rho=0.3434\) \\
RobotWin & demo & 5 seeds & Pi0.5 vision & 0.8715 & 0.8452--0.8916 \\
RobotWin & demo & 5 seeds & Pi0.5 VLM final & 0.8407 & 0.8219--0.8555 \\
RobotWin & demo & 5 seeds & Pi0.5 layer 08 & 0.8276 & 0.8005--0.8362 \\
RobotWin & task & 5 seeds & Pi0.5 vision & -0.8528 & \(\rho=0.1088\) \\
RobotWin & task & 5 seeds & Pi0.5 VLM final & -0.9314 & \(\rho=0.1848\) \\
RobotWin & task & 5 seeds & Pi0.5 layer 08 & -0.7790 & \(\rho=0.1330\) \\
\bottomrule
\end{tabular}
\caption{Cross-benchmark probing checks and task-split boundaries. Extra statistic reports Spearman \(\rho\) for CALVIN-D and RobotWin task rows, and the seed range for RobotWin demo rows.}
\label{tab:appendix-cross-benchmark}
\end{table}

\paragraph{Supporting online runs and sampling regimes.}
Table~\ref{tab:appendix-supporting-online} reports online runs that informed the selection design but are not pooled with the balanced production estimates in Table~\ref{tab:online-positive}.  The first rows show that positive value-guided directions appeared before the formal merged runs, including on Pi0.5 and OpenVLA.  The \(K\)- and temperature-sweep rows show that the selection gain depends on the candidate distribution: the probe has leverage only when the sampled prefixes expose alternatives that can be separated by the downstream score.  For this reason, the main text treats open middle drawer as a boundary case rather than folding the early positive drawer pilot into the formal estimate.

\begin{table}[!htbp]
\centering
\setlength{\tabcolsep}{2pt}
\begin{tabular}{@{}
  >{\raggedright\arraybackslash}p{\dimexpr 0.2\linewidth - 1.72\tabcolsep\relax}
  >{\raggedright\arraybackslash}p{\dimexpr 0.25\linewidth - 1.72\tabcolsep\relax}
  >{\centering\arraybackslash}p{\dimexpr 0.09\linewidth - 1.72\tabcolsep\relax}
  >{\centering\arraybackslash}p{\dimexpr 0.1\linewidth - 1.72\tabcolsep\relax}
  >{\raggedleft\arraybackslash}p{\dimexpr 0.1\linewidth - 1.72\tabcolsep\relax}
  >{\raggedleft\arraybackslash}p{\dimexpr 0.13\linewidth - 1.72\tabcolsep\relax}
  >{\raggedleft\arraybackslash}p{\dimexpr 0.13\linewidth - 1.72\tabcolsep\relax}
@{}}
\toprule
Family & Task / setting & \(n\) & Random & Value & \(\Delta\) V--R & Status \\
\midrule
Pi0.5 full & Open middle drawer & 50 & 32\% & 48\% & +16 pp & support \\
Pi0.5 full & Push plate & 50 & 32\% & 46\% & +14 pp & support \\
Pi0.5 full & Wine rack & 50 & 40\% & 48\% & +8 pp & support \\
OpenVLA 50ep & Open middle drawer & 50 & 38\% & 70\% & +32 pp & support \\
OpenVLA 50ep & Open top drawer & 50 & 40\% & 52\% & +12 pp & support \\
\(K\) sweep & Push plate, \(K=8\) & 50 & 32\% & 42\% & +10 pp & positive \\
\(K\) sweep & Wine rack, \(K=8\) & 50 & 30\% & 34\% & +4 pp & positive \\
\(K\) sweep & Push plate, \(K=32\) & 50 & 34\% & 48\% & +14 pp & positive \\
\(K\) sweep & Wine rack, \(K=32\) & 50 & 46\% & 44\% & -2 pp & mixed \\
Temperature & Open middle, \(t=0.7\) & 20 & 25\% & 90\% & +65 pp & positive \\
Temperature & Open middle, \(t=0.3\) & 20 & 65\% & 55\% & -10 pp & mixed \\
Temperature & Open top, \(t=0.3\) & 20 & 15\% & 30\% & +15 pp & positive \\
Temperature & Open top, \(t=0.7\) & 20 & 55\% & 55\% & 0 pp & mixed \\
\bottomrule
\end{tabular}
\caption{Supporting online and sampling-regime results. These runs show positive or mixed directions but are not pooled with the formal Pi0.5 estimates in Table~\ref{tab:online-positive}.}
\label{tab:appendix-supporting-online}
\end{table}

\paragraph{Early candidate-set sweep.}
The \(K=16\) candidate-set size was fixed before the balanced production runs.  Table~\ref{tab:appendix-k-sweep} reports the early two-task sweep behind that choice.  Increasing \(K\) from 4 or 8 to 16 improved overall success from 60\% to 80\%, while also increasing wall time substantially.  We use this result only as design calibration for the later selector; it is not pooled with the formal online estimates or the later \(K=8/K=32\) ablations.

\begin{table}[!htbp]
\centering
\setlength{\tabcolsep}{4pt}
\begin{tabular}{@{}r c c c@{}}
\toprule
\(K\) & Overall success & Mean steps & Mean wall time \\
\midrule
4 & 60.00\% & 216.0 & 6.1 min \\
8 & 60.00\% & 212.1 & 11.4 min \\
16 & 80.00\% & 182.3 & 26.7 min \\
\bottomrule
\end{tabular}
\caption{Early two-task candidate-set sweep used to choose \(K=16\) for the main simulator-backed selector.}
\label{tab:appendix-k-sweep}
\end{table}

\paragraph{Compute cost of controlled selection.}
\label{sec:appendix-compute}
Table~\ref{tab:appendix-compute} gives the wall-clock accounting for the formal Pi0.5 online payloads.  Random selection and value-guided selection both operate over the same candidate-generation regime, but value-guided selection additionally restores simulator states, rolls out candidates, extracts frozen features, and scores each candidate with the probe.  The resulting cost is roughly \(2\times\) random selection and more than an order of magnitude above greedy decoding.  The numbers quantify the controlled intervention: they measure what is paid to make candidate futures observable and test whether the representation-derived score can improve the committed action.

\begin{table}[!htbp]
\centering
\setlength{\tabcolsep}{2pt}
\begin{tabular}{@{}
  >{\raggedright\arraybackslash}p{\dimexpr 0.21\linewidth - 1.67\tabcolsep\relax}
  >{\centering\arraybackslash}p{\dimexpr 0.06\linewidth - 1.67\tabcolsep\relax}
  >{\centering\arraybackslash}p{\dimexpr 0.17\linewidth - 1.67\tabcolsep\relax}
  >{\centering\arraybackslash}p{\dimexpr 0.18\linewidth - 1.67\tabcolsep\relax}
  >{\centering\arraybackslash}p{\dimexpr 0.18\linewidth - 1.67\tabcolsep\relax}
  >{\centering\arraybackslash}p{\dimexpr 0.2\linewidth - 1.67\tabcolsep\relax}
@{}}
\toprule
Task & \(n\) & Greedy & Random & Value-guided & Value/random wall \\
\midrule
Open middle drawer & 150 & 38.67\%, 19.5s & 37.33\%, 212.2s & 39.33\%, 434.3s & 2.05\(\times\) \\
Push plate         & 300 & 26.67\%, 21.7s & 33.67\%, 229.8s & 44.33\%, 481.7s & 2.10\(\times\) \\
Wine rack          & 300 & 35.67\%, 20.5s & 36.00\%, 224.8s & 44.00\%, 451.8s & 2.01\(\times\) \\
\bottomrule
\end{tabular}
\caption{Compute diagnostics for the formal Pi0.5 online payloads. Success rates are paired with mean wall time per episode in seconds.}
\label{tab:appendix-compute}
\end{table}

\paragraph{Goal-swap diagnostics.}
\label{sec:appendix-goal-swap}
Table~\ref{tab:appendix-goal-swap} separates two questions that are easy to conflate.  The vision-only control checks the mechanics of the swap: when the goal pathway is bypassed, changing the goal text produces zero feature movement and zero score movement.  The VLM rows then show that Pi0.5 language-conditioned features do move under goal swaps, and that probe scores move with them.  The sign is not stable across seeds and layers, so these rows should not be read as a calibrated reward-model test.  They support a narrower claim: the probe is computed from a representation whose features are affected by the language goal, rather than from a purely visual statistic.


\begin{table}[!htbp]
\centering
\setlength{\tabcolsep}{2pt}
\begin{tabular}{@{}
  >{\raggedright\arraybackslash}p{\dimexpr 0.26\linewidth - 1.72\tabcolsep\relax}
  >{\centering\arraybackslash}p{\dimexpr 0.1\linewidth - 1.72\tabcolsep\relax}
  >{\centering\arraybackslash}p{\dimexpr 0.12\linewidth - 1.72\tabcolsep\relax}
  >{\centering\arraybackslash}p{\dimexpr 0.14\linewidth - 1.72\tabcolsep\relax}
  >{\centering\arraybackslash}p{\dimexpr 0.14\linewidth - 1.72\tabcolsep\relax}
  >{\centering\arraybackslash}p{\dimexpr 0.12\linewidth - 1.72\tabcolsep\relax}
  >{\raggedleft\arraybackslash}p{\dimexpr 0.12\linewidth - 1.72\tabcolsep\relax}
@{}}
\toprule
Probe / run family & Runs & Pairs & Score \(\Delta\) & Pos. frac. & Feature \(L_2\) & Cosine \\
\midrule
Vision encoder control & 16 & 14,080 & 0.0000 & 0.0000 & 0.0000 & 1.0000 \\
VLM final, seed 7 & 16 & 14,080 & 0.1226 & 0.5366 & 1.9646 & 0.9985 \\
VLM layer 08, seed 7 & 8 & 7,040 & -0.0191 & 0.3706 & 15.2770 & 0.9999 \\
VLM final, seed 17 & 6 & 5,280 & -0.0793 & 0.4545 & 1.9412 & 0.9986 \\
\bottomrule
\end{tabular}
\caption{Goal-swap diagnostics. Vision-only rows are negative controls; VLM rows test whether language-conditioned features and probe scores move when the goal text is swapped.}
\label{tab:appendix-goal-swap}
\end{table}

\paragraph{Per-task goal-swap direction.}
Averaging over goal swaps hides a structured asymmetry across tasks.  Table~\ref{tab:appendix-goal-swap-tasks} breaks out the seed-7 \texttt{vlm\_final\_output} run by task.  Drawer-style tasks move in the negative direction under the original-minus-swapped score convention, whereas rack, container, and stove-related tasks tend to move positive.  This is the basis for the interpretation in Section~\ref{sec:goal-conditioning}: the probe is goal-conditioned, but its score is not a calibrated reward that always assigns a higher value to the matched goal.

\begin{table}[!htbp]
\centering
\setlength{\tabcolsep}{4pt}
\begin{tabular}{@{}p{0.28\linewidth}c c@{}}
\toprule
Task & Score \(\Delta\) & Pos. frac. \\
\midrule
Open middle drawer & -1.0702 & 0.1648 \\
Open top drawer & -1.4007 & 0.0920 \\
Push plate & -0.1548 & 0.4091 \\
Bowl on plate & 0.4878 & 0.6477 \\
Bowl on stove & 0.1896 & 0.5148 \\
Bowl on cabinet & 0.3079 & 0.5659 \\
Cream cheese in bowl & 0.9177 & 0.8125 \\
Wine rack & 1.2425 & 0.9080 \\
Wine on cabinet & 1.0422 & 0.8477 \\
Turn on stove & 0.5568 & 0.6716 \\
\bottomrule
\end{tabular}
\caption{Per-task goal-swap direction for the seed-7 \texttt{vlm\_final\_output} probe.  Score \(\Delta\) is original-goal score minus swapped-goal score.}
\label{tab:appendix-goal-swap-tasks}
\end{table}

\paragraph{Teacher-student amortization.}
Table~\ref{tab:appendix-student} reports a small attempt to amortize the simulator-backed teacher into a cheaper student scorer.  The LIBERO top-drawer pilot is positive, improving from 35/50 to 43/50 successes, which shows that teacher preferences can in some cases be converted into a direct online scorer.  The middle-drawer and RobotWin rows are weaker: the LIBERO middle-drawer student does not improve over random selection, and the RobotWin validation scorer remains close to chance.  We therefore treat the student experiment as evidence that amortization is a plausible extension, not as a replacement for the teacher selector used in the main results.

\begin{table}[!htbp]
\centering
\setlength{\tabcolsep}{2pt}
\begin{tabular}{@{}
  >{\raggedright\arraybackslash}p{\dimexpr 0.29\linewidth - 1.5\tabcolsep\relax}
  >{\raggedright\arraybackslash}p{\dimexpr 0.25\linewidth - 1.5\tabcolsep\relax}
  >{\raggedright\arraybackslash}p{\dimexpr 0.21\linewidth - 1.5\tabcolsep\relax}
  >{\centering\arraybackslash}p{\dimexpr 0.25\linewidth - 1.5\tabcolsep\relax}
@{}}
\toprule
Setting & Baseline / data & Student / output & Interpretation \\
\midrule
LIBERO top-drawer       & 50 episodes/strategy; random 35/50 (70\%) & student 43/50 (86\%)          & +16 pp pilot gain \\
LIBERO middle drawer    & 50 episodes/strategy; random 25/50 (50\%) & student 23/50 (46\%)          & weak case \\
RobotWin teacher replay & 346 records; 6 tasks                      & VLM final features            & teacher data collected \\
RobotWin student ranker & validation set                            & pairwise 0.5244; top-1 0.1461 & weak scorer \\
\bottomrule
\end{tabular}
\caption{Teacher-student extension diagnostics. The LIBERO top-drawer result is the positive pilot; the remaining rows show weaker student settings or data-collection status.}
\label{tab:appendix-student}
\end{table}



\end{document}